\documentclass[final,times,twocolumn]{elsarticle}

\usepackage{amssymb}

\usepackage{times}
\usepackage{soul}
\usepackage{url}
\usepackage[hidelinks]{hyperref}
\usepackage[utf8]{inputenc}
\usepackage[small]{caption}
\usepackage{graphicx}
\usepackage{amsmath}
\usepackage{amsthm}
\usepackage{amsfonts}
\usepackage{booktabs}
\usepackage{algorithm}
\usepackage{algorithmic}
\usepackage[switch]{lineno}
\usepackage{cleveref}
\usepackage{xcolor}
\usepackage{comment}
\usepackage{natbib}
\usepackage[utf8]{inputenc}
\usepackage{booktabs} 

\journal{Neurocomputing}

\usepackage{lipsum}

\begin{document}

\newcommand\blfootnote[1]{%
\begingroup 
\renewcommand\thefootnote{}\footnote{#1}%
\addtocounter{footnote}{-1}%
\endgroup 
}

\begin{frontmatter}



\title{Multi-Agent, Human-Agent and Beyond: A Survey on Cooperation in Social Dilemmas}

\author[inst1]{Chunjiang Mu$^{\star}$}
\author[inst2]{Hao Guo$^{\star}$}
\author[inst3]{Yang Chen}
\author[inst4]{Chen Shen}
\author[inst1,inst5]{Die Hu}
\author[inst6]{Shuyue Hu\corref{cor}}
\ead{hushuyue@pjlab.org.cn}
\author[inst1]{Zhen Wang\corref{cor}}
\ead{w-zhen@nwpu.edu.cn}

\tnotetext[equal]{Equal contributions.}
\cortext[cor]{Corresponding authors.}


\affiliation[inst1]{
    organization={Northwestern Polytechnical University},
    city={Xi'an},
    postcode={710072}, 
    country={China}
}

\affiliation[inst2]{
    organization={Tsinghua University},
    city={Beijing},
    postcode={100084},
    country={China}
}
            
\affiliation[inst3]{
    organization={University of Auckland},
    city={Auckland},
    postcode={1001}, 
    country={New Zealand}
}

\affiliation[inst4]{
    organization={Kyushu University},
    city={Fukuoka},
    postcode={816-8580}, 
    country={Japan}
}

\affiliation[inst5]{
    organization={The Hong Kong Polytechnic University},
    city={Hong Kong},
    country={China}
}

\affiliation[inst6]{
    organization={Shanghai Artificial Intelligence Laboratory},
    city={Shanghai},
    postcode={200030},
    country={China}
}

\begin{abstract}
The study of cooperation within social dilemmas has long been a fundamental topic across various disciplines, including computer science and social science. Recent advancements in Artificial Intelligence (AI) have significantly reshaped this field, offering fresh insights into understanding and enhancing cooperation. This survey examines three key areas at the intersection of AI and cooperation in social dilemmas. First, focusing on multi-agent cooperation,  we review the intrinsic and external motivations that support cooperation among rational agents, and the methods employed to develop effective strategies against diverse opponents. Second, looking into human-agent cooperation, we discuss the current AI algorithms for cooperating with humans and the human biases towards AI agents. Third, we review the emergent field of leveraging AI agents to enhance cooperation among humans. We conclude by discussing future research avenues, such as using large language models, establishing unified theoretical frameworks, revisiting existing theories of human cooperation, and exploring multiple real-world applications.
\end{abstract}




\end{frontmatter}


\section{Introduction}

Social dilemmas (SDs, e.g., prisoner's dilemma), spanning various domains including environmental pollution, public health crises, and resource management, present a fundamental conflict between personal interests and the common good~\citep{nowak2006five}.
While cooperation is beneficial for the collective, individuals are tempted to exploit or free-ride others' efforts, potentially leading to a tragedy of the commons.
Historically rooted in the study of biological altruism~\citep{smith1982evolution}, the traditional research on cooperation in SDs has unveiled the pivotal roles of reciprocity, social preferences, and communications in fostering cooperative behaviors in human-human interactions~\citep{fehr2002strong, li2018punishment, rand2013human, wang2020communicating}.
Recently, propelled by advances in artificial intelligence (AI), this field has been undergoing a profound transformation---as AI agents now increasingly represent and engage with humans, our understanding of how cooperation emerges, evolves, and sustains in SDs is being significantly reshaped.
This is particularly evident in two lines of research: \emph{multi-agent cooperation}, where AI agents interact with each other in SDs, and \emph{human-agent cooperation}, which examines the intricacies of human interactions with AI agents in SDs. 

\begin{figure*}[h]
    \centering
    \includegraphics[width=1\textwidth]{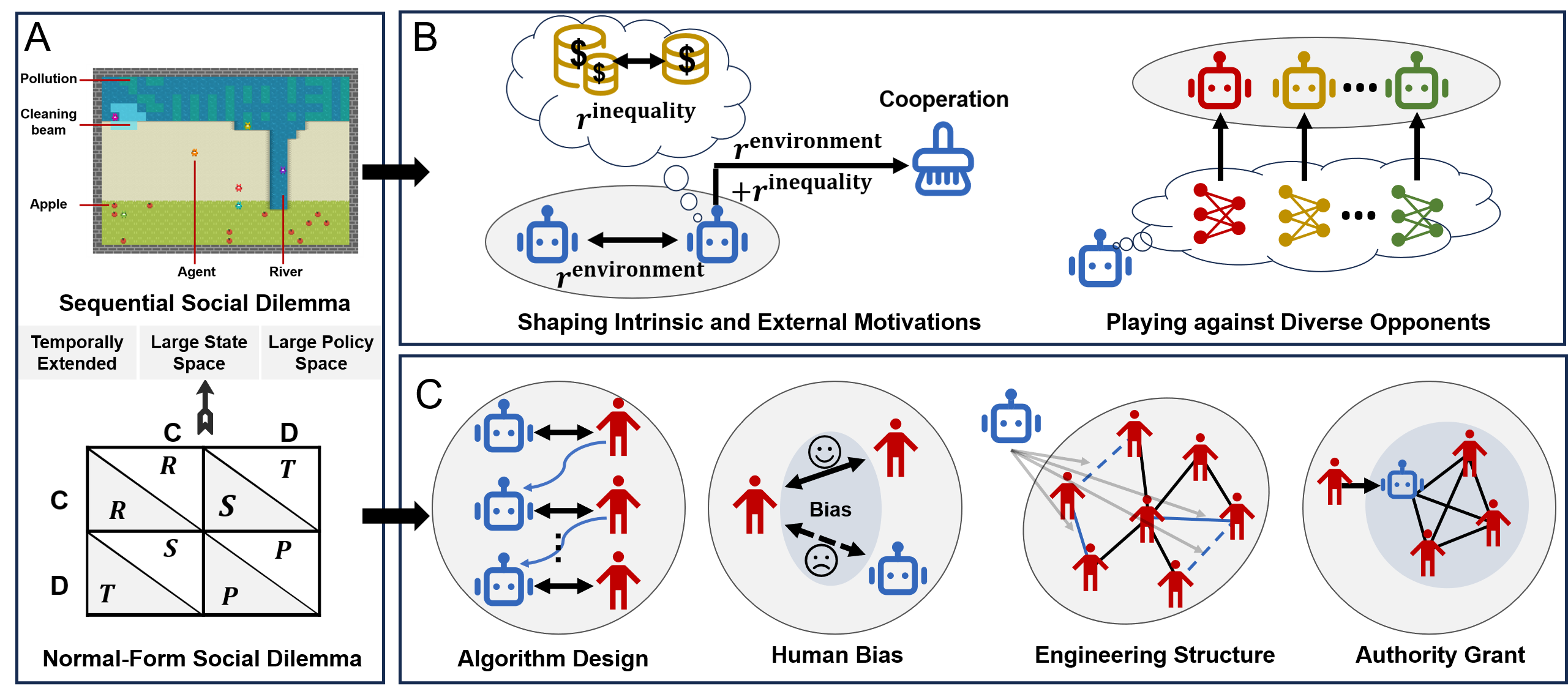}
    \caption{Understanding cooperation within multi-agent, human-agent systems, and beyond. (A) Normal-form social dilemmas and sequential social dilemmas. (B) Two approaches to solving sequential social dilemmas in multi-agent systems: i) promoting cooperation among agents through shaping their intrinsic and external motivations; ii) devising and selecting strategies in response to diverse opponents. (C) Four perspectives of studying cooperation in human-agent hybrid systems: i) designing algorithms for cooperating with humans; ii) identifying and mitigating human biases in human-agent cooperation; iii) scaffolding cooperation in human-human interactions, e.g., by engineering the interaction structure; iv) delegating human decision making to agents.}
    \label{fig:SSD problem}
\end{figure*}

A fundamental assumption of AI agents is their rationality. 
This raises a critical question: how can rational agents be steered towards effective cooperation, requiring them to overcome the lure of exploiting or free-riding others? 
This question has received much interest in recent studies of multi-agent cooperation, particularly in the context of sequential social dilemmas (SSDs) \citep{leibo2017multi}.
As opposed to normal-form SDs, SSDs are characterized by stochastic environments as well as larger state and policy spaces. 
It is shown that embedding human-like motives, such as fairness \citep{hughes2018inequity} and social preferences \citep{mckee2020social}, into agents' rewards can promote cooperative behavior. In addition, mechanisms, like peer rewarding \citep{wang2018exploiting} and formal contracts \citep{christoffersen2023get}, have also shown potential in fostering cooperation among agents. On the other hand, it is shown that 
agents can safeguard themselves against other agents' exploitation by pre-training a suite of policies and then choosing them adaptively in real-time \citep{lerer2017maintaining}, or by influencing the future policies of their opponents \citep{foerster2018learning}. 

In addition to multi-agent cooperation, human-agent cooperation is equally important, as interactions between humans and agents are becoming ubiquitous. 
This field is centered around two key questions.
The first concerns the design of AI algorithms: how can  
agents cooperate with humans at a level comparable to human-human cooperation? Recent studies have shown that combining expert algorithms and reinforcement learning can lead to human-level cooperation in SDs involving both humans and agents~\citep{crandall2018cooperating}.  %
Moreover, simpler strategies, such as the extortion and generous strategies, are also instrumental in enhancing human-agent cooperation~\citep{hilbe2014extortion}. 
The second question delves into the socio-cognitive aspect: in what ways do human perceptions and reactions differ when dealing with AI agents compared to other humans? 
Interestingly, behavioral evidence has revealed that humans are prone to cooperate less and even exploit AI agents more when they are aware of the agents' non-human nature~\citep{ishowo2019behavioural}. 
To mitigate such human bias towards agents, one approach is to embed culturally relevant signals and emotionally expressive characteristics in AI agents~\citep{de2019cooperation}.

Beyond multi-agent and human-agent cooperation, emerging research indicates that AI can enhance human-human cooperation in SDs, consequently, in turn, enriching the long-standing study of cooperative behaviors in human societies.
 Specifically, AI agents have been found to promote cooperation among humans by providing recommendations on partner selection~\citep{mckee2023scaffolding}. 
Moreover, analyses based on evolutionary game theory suggest that human-AI hybrid systems can surpass pure human populations in achieving higher cooperation levels~\citep{guoFacilitatingCooperationHumanagent2023}. 

Each aforementioned field---multi-agent cooperation, human-agent cooperation, as well as the emergent field of leveraging AI agents to promote human-human cooperation---has attracted significant interest though, there lacks a comprehensive review that integrates insights from these fields.
Existing reviews have touched on learning in SSDs, but those reviews primarily emphasize multi-agent reinforcement learning (MARL), with SSDs as specialized scenarios within mixed-motive MARL ~\citep[e.g.,][]{zhang2021multi}, without sufficiently addressing the close connection between SDs and human cooperation.
On the other hand, numerous reviews have examined cooperative behaviors but predominantly within the context of human societies, and are restricted to normal-form SDs ~\citep[e.g.,][]{rand2013human, perc2017statistical}, without fully acknowledging the increasingly significant role of AI in the study of SDs. In light of these observations, 
this survey aims to provide a comprehensive understanding that unifies these fields, and encapsulates the diverse facets of cooperation across these interconnected fields, as illustrated in Figure 1. 
This can not only reflect the current state of research at the intersection of AI and cooperation in SDs, but also illuminate potential paths for future investigations, particularly in the interplay of multi-agent, human-agent, and human-human cooperation. 
As AI becomes more integrated into human society, we call for increased research in the field of cooperative AI \citep{dafoe2021cooperative}.

The remainder of this survey is organized as follows.
 We begin by briefly introducing the normal-form SDs and SSDs. We then delve into multi-agent cooperation, summarizing the mechanisms that support agents to achieve mutual cooperation and the methods for developing strategies against diverse opponents. The focus then shifts to human-agent cooperation, discussing the current AI algorithms for cooperating with humans and the human biases towards AI agents. Subsequently, we review how the advances of AI agents can inspire human-human cooperation. Finally, we conclude with discussions on future research directions: (i) enhancing the study of cooperation with large language models, (ii) establishing theoretical frameworks for cooperation in  SSDs and human-agent cooperation, (iii) applications to multiple real-world scenarios, (iv) bridging human-agent cooperation and SSDs, and (v) revisiting the existing theory of cooperation in human societies.

\section{Preliminaries}
This section defines normal-form and sequential social dilemmas, upon which the multi-agent and human-agent cooperation methods surveyed in this paper are built.
\subsection{Normal-Form Social Dilemmas}
Social dilemmas (SDs) involve a conflict between immediate self-interest and longer-term collective interests \citep{van2013psychology}. 
A normal-form SD is a general-sum normal-form game, where both players must choose one of two strategies simultaneously: cooperation ($C$) or defection ($D$).
Mutual cooperation yields the reward $R$ for both, mutual defection leads to the punishment $P$ for both, and different choices give the cooperator the sucker’s payoff $S$ and the defector the temptation payoff $T$. 
In a SD, these payoffs satisfy the following relationships:
\begin{itemize}
    \item $R > P$: mutual cooperation values higher than mutual defection {\em and}
    \item $R > S$: mutual cooperation values higher unilateral cooperation {\em and}
    \item $2R > T+S$: social welfare of mutual cooperation values higher than that of unilateral cooperation {\em and}
    \item $T > R$: unilateral defection for greed values higher than mutual cooperation {\em or} 
    \item $P > S$:  unilateral defection for fear values higher than mutual cooperation.
\end{itemize}
The utility functions of SD can be typically represented by the following payoff matrix $M$:

\begin{equation*}
\begin{array}{c|cc}
 & C & D \\
\hline
C &R & S \\
D &T & P \\
\end{array}.
\end{equation*}
The elements in the matrix are the payoffs for the row player, and the payoffs for the column player are at the transposed mirror position of the row player.
Next, we introduce three classic models of SDs: Prisoner’s Dilemma (PD), Stag Hunt (SH) and Hawk-Dove (HD).
\paragraph{Prisoner’s Dilemma}
The PD is a classic social dilemma where cooperation results in the highest collective payoff, yet individuals acting in their own best interest tend to defect, leading to poorer outcomes for both involved players \citep{tucker1950two}.
The payoff rankings for a PD specifically are  $T>R>P>S$.
An example of a payoff matrix for this scenario is presented below:
\begin{equation*}
\begin{array}{c|cc}
\text{PD}& C & D \\
\hline
C & 3 & 0 \\
D & 4 & 1 \\
\end{array}.
\end{equation*}
There is unique pure strategy Nash equilibrium \footnote{A Nash equilibrium in a game is a strategy profile from which no player can increase their payoff by unilaterally changing their strategy, provided the strategies of the other players remain unchanged \citep{RePEc:mtp:titles:0262650401}.} $(D, D)$ in a PD.

\paragraph{Stag Hunt}
In an SH, the highest collective payoff is achieved through mutual cooperation; however, players are tempted by lower-risk individual strategies due to fear of defection. The payoff structure adheres to the relationship $R>T\geq P>S$, indicating that defection may occur out of fear. Below is an illustrative payoff matrix for the SH:
\begin{equation*}
\begin{array}{c|cc}
\text{SH} & C & D \\
\hline
C &5 & 0 \\
D &2 & 1 \\
\end{array}.
\end{equation*}
There are two pure strategy Nash equilibrium $(C,C)$ and $(D, D)$ in an SH.
Additionally, the SH also have a mixed strategy equilibrium, depending on the specific setup of the payoffs.

\paragraph{Hawk-Dove}
The HD involves strategies of cooperation and defection in the context of resource competition among animals or individuals \citep{RePEc:mtp:titles:0262650401}. 
It reflects the trade-offs between risk and reward. 
The payoffs typically follow the order $T>R>S>P$, suggesting that players may choose aggression driven by greed. 
Below is an example of a payoff matrix for the HD:
\begin{equation*}
\begin{array}{c|cc}
\text{HD} & C&D \\
\hline
C&3 & 1 \\
D & 4 & 0\\
\end{array}.
\end{equation*}

There are two pure strategy Nash equilibrium $(C, D)$ and $(D, C)$ as well as a mixed strategy Nash equilibrium.
Another two game models mathematically equivalent to the HD are Chicken game and Snowdrift game.

\subsection{Sequential Social Dilemmas}
Sequential Social Dilemmas (SSDs) are social dilemmas happening over a sequence of time steps and defined in stochastic games that consider the states of the decision-making environment. 
Formally, an $n$-agent SSD is a tuple $\langle \mathcal{M},\varPi = \varPi_c\bigsqcup  \varPi_d \rangle$ \citep{leibo2017multi}, where
$\mathcal{M}$ 
is a stochastic game (also known as Markov game), $\varPi_c$ is a set of cooperative policies, and $\varPi_d$ is a  set of defective policies.
A stochastic game is a repeated game associated with probabilistic state transitions, where each agent takes action based on its policy 
simultaneously at the current environmental state; then the environment will transform to a new state, and each agent will receive a reward from the environment as a result of agents' actions and the state transition. 

For an SSD, $\varPi_c$ and $\varPi_d$ induce the following manners.
Consider a set $N_c$ of agents where each agent $j \in N_c$ adopts a policy $\pi^j \in \varPi_c$, and a set $N_d$ of agents where each agent $k \in N_d$ adopts a policy $\pi^k \in \varPi_d$, such that $N_c \cup N_d =N, N_c \cap  N_d =\emptyset$.
Denote the expected reward of the agent $i \in N$ by $R^i(l)$ in an episode of the stochastic game, with the number of agents adopting cooperative policies being $l=|N_c|$. 
\citet{hughes2018inequity} propose conditions for an $n$-agent SSD to hold:

\begin{itemize}
    \item $R^j(|N|)>R^k(0),\forall{j\in N_c}\ \forall{k\in N_d}$: cooperators under full cooperation gain more than defectors under full defection {\em and}
    \item $R^j(|N|)>R^j(1), \forall{j\in N_c}$: cooperators under full cooperation gain more than one cooperator under the others' defection {\em and}
    \item$R^k(l)>R^j(l), \forall{k\in N_d}\ \forall{j\in N_c}$ if $|N_c|\ge|N_d|$: defectors gain more than cooperators when the number of cooperators is relatively large (for fear) {\em or}
    \item $R^k(l)>R^j(l), \forall{k\in N_d}\ \forall{j\in N_c}$ if $|N_c|\leq|N_d|$: defectors gain more than cooperators when the number of cooperators is relatively small (for greed).
\end{itemize}

Compared to the highly abstract normal-form SDs, SSDs have some characteristics that are more aligned with real-world scenarios, which are well-summarized by \citet{leibo2017multi}:
First, real-world social dilemmas are temporally extended.
Then, from an individual's policy perspective, cooperation and defection are labels that apply to policies implementing individual strategic decisions, and one's cooperativeness may be a graded quantity.
Additionally, from the perspective of  game participation, decisions to cooperate or defect occur only quasi-simultaneously, and sometimes decisions must be made when only having partial information about the state of the world and other players' activities.

Next we introduce several commonly used environments of SSDs.

\begin{figure}[h]
    \centering
    \includegraphics[width=0.3\textwidth]{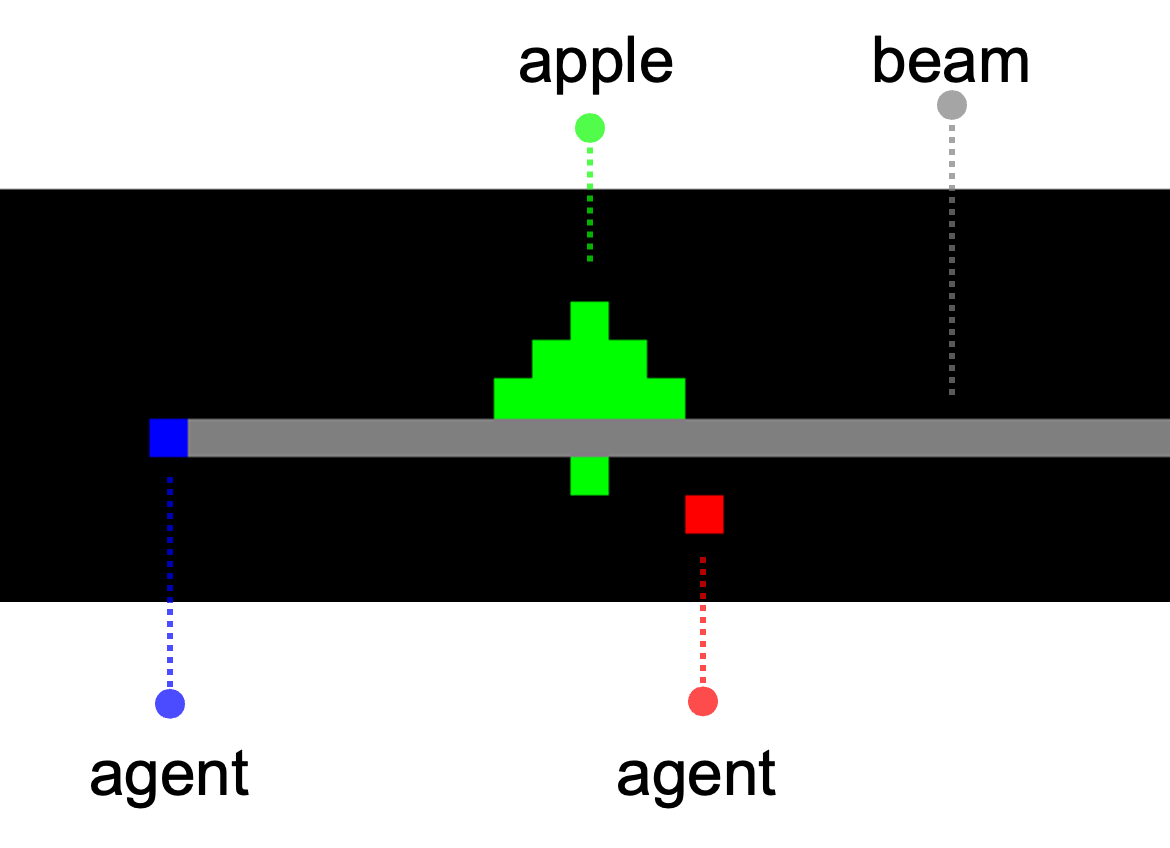}
    \caption{Illustration of Gathering \citep{leibo2017multi}.}
    \label{fig:gathering}
\end{figure}

\paragraph{Gathering \textup{\citep{leibo2017multi}}}
Two agents aim to accumulate apples, represented by green pixels on the map. 
An agent earns a reward of 1 for each apple collected, after which the apple is temporarily removed and respawns after $N_\text{apple}$ frames. 
Agents are equipped with the capability to project a beam in alignment with their orientation. An agent that is struck by the beam twice is ``tagged'' and consequently removed from play for $N_\text{tagged}$ frames. 
Importantly, tagging does not confer rewards; its sole purpose may lie in reducing competition for apple resources.
The illustration of Gathering is shown in \Cref{fig:gathering}.

\begin{figure}[h]
    \centering
    \includegraphics[width=0.35\textwidth]{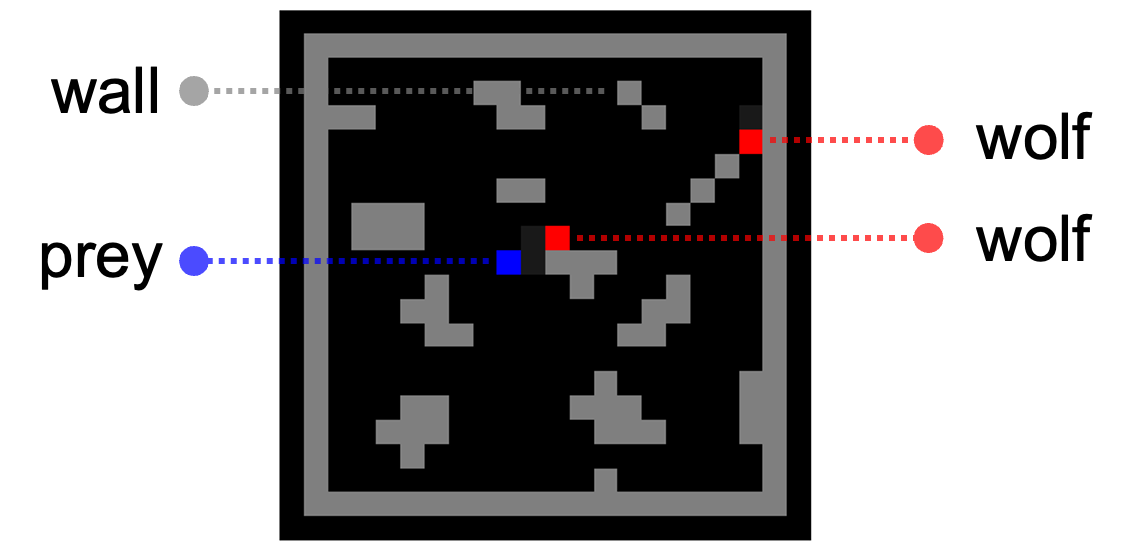}
    \caption{Illustration of Wolfpack \citep{leibo2017multi}.}
    \label{fig:wolfpack}
\end{figure}

\paragraph{Wolfpack \textup{\citep{leibo2017multi}}}
Two agents (wolves) aim to chase a third agent (the prey). 
When either wolf touches the prey, all wolves within the capture radius receive a reward. 
The reward received by the capturing wolves is proportional to the number of wolves in the capture radius. 
The idea is that a lone wolf can capture the prey, but is at risk of losing the carcass to scavengers. 
However, when the two wolves capture the prey together, they can better protect the carcass from scavengers and hence receive a higher reward. 
A lone-wolf capture provides a reward of $r_\text{lone}$ and a capture involving both wolves is worth $r_\text{team}$.
The illustration of Wolfpack is shown in \Cref{fig:wolfpack}.

\begin{figure}[h]
    \centering
    \includegraphics[width=0.3\textwidth]{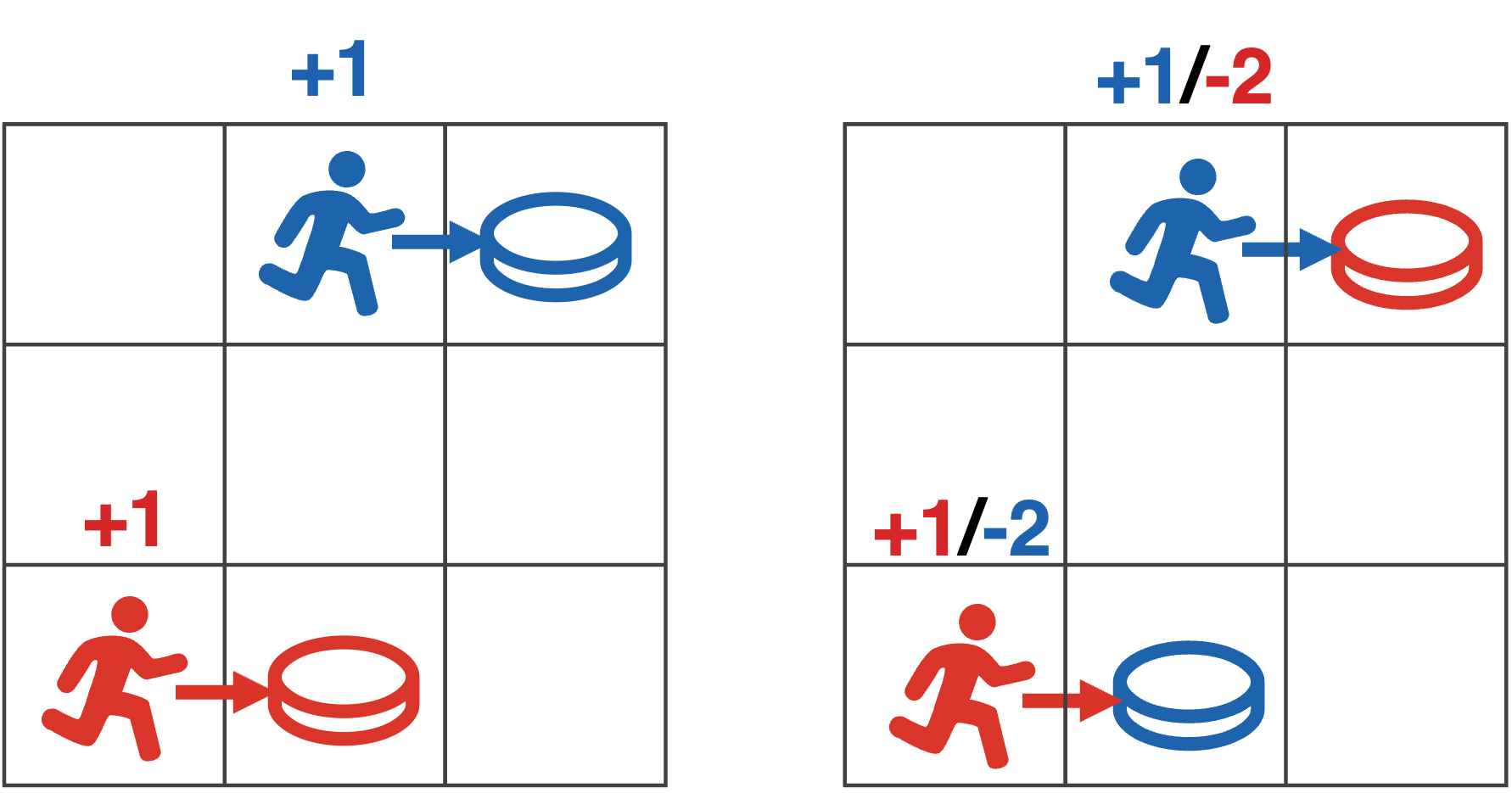}
    \caption{Illustration of Coins \citep{lerer2017maintaining}.}
    \label{fig:coingame}
\end{figure}

\begin{figure}[h]
    \centering
    \includegraphics[width=0.35\textwidth]{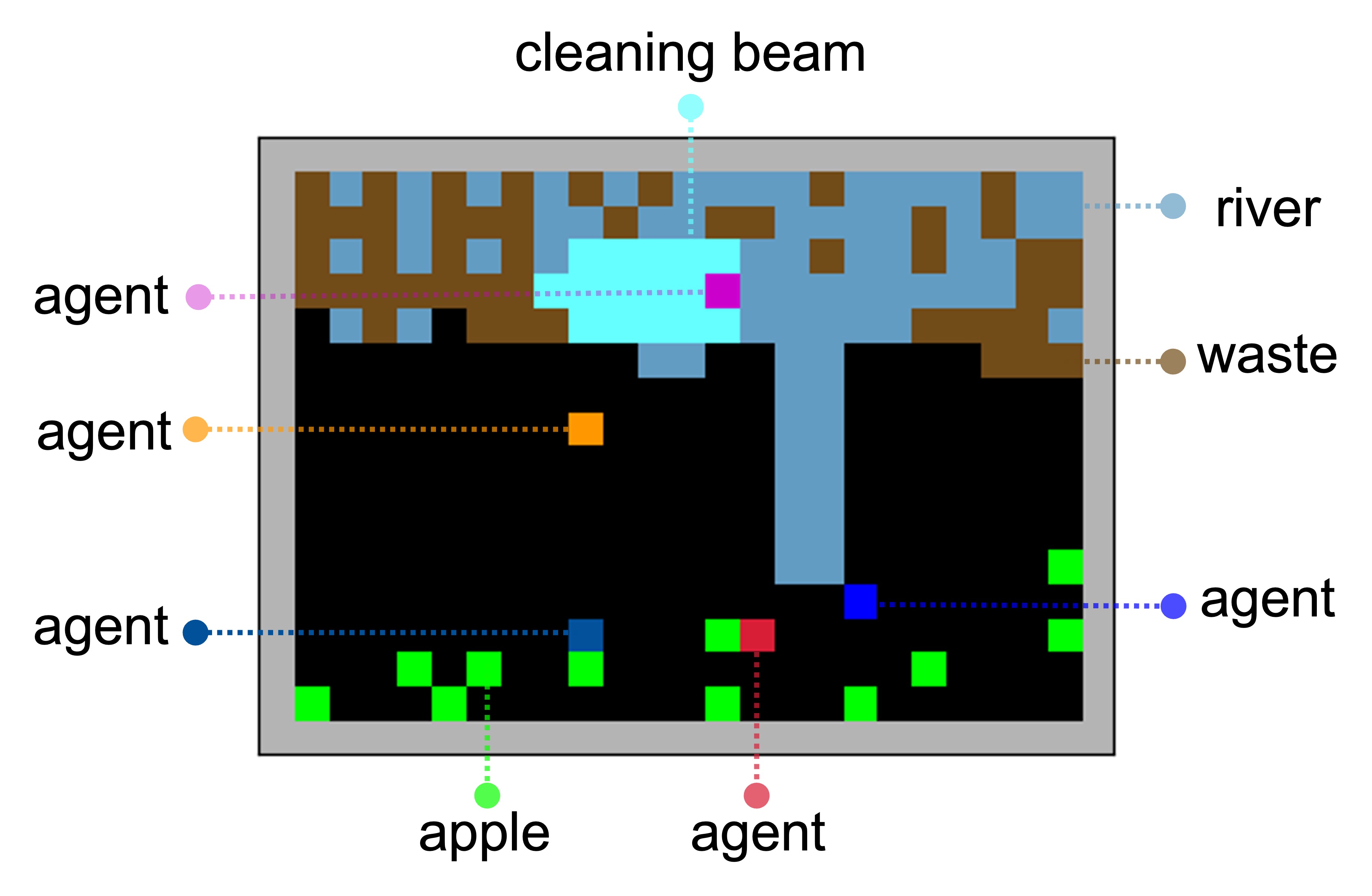}
    \caption{Illustration of CleanUp \citep{hughes2018inequity}.}
    \label{fig:cleanup}
\end{figure}

\paragraph{Coins \textup{\citep{lerer2017maintaining}}}
Two agents labeled in red and blue respectively, are tasked with picking up coins, also labeled red and blue respectively.
If an agent picks up any coin by moving into the same position as the coin, it will earn a reward of 1. 
However, if the coin it pick up is in color of the other agent, the other agent receives a reward of -2. 
Thus, if both agents play greedily and pick up every coin, the expected reward for both agents is 0.
The illustration of Coins is shown in \Cref{fig:coingame}.
\paragraph{CleanUp \textup{\citep{hughes2018inequity}}}
The goal of all agents is to collect apples from a ﬁeld. 
Each apple provides a reward of 1. 
The spawning of apples is controlled by a geographically separate aquifer that supplies water and nutrients. 
The river ﬁlls up with waste over time, lowering the respawn rate of apples linearly. 
For sufﬁciently high waste levels, no apples can spawn. 
To let apples to spawn, agents must clean waste in the river by firing clean beam.
Furthermore, agents can also fire penalty beams, inflicting a reward of -50 reward on the hit agent at the cost of a -1 reward to themselves.
The illustration of CleanUp is shown in \Cref{fig:cleanup}.

\begin{figure}[h]
    \centering
    \includegraphics[width=0.43\textwidth]{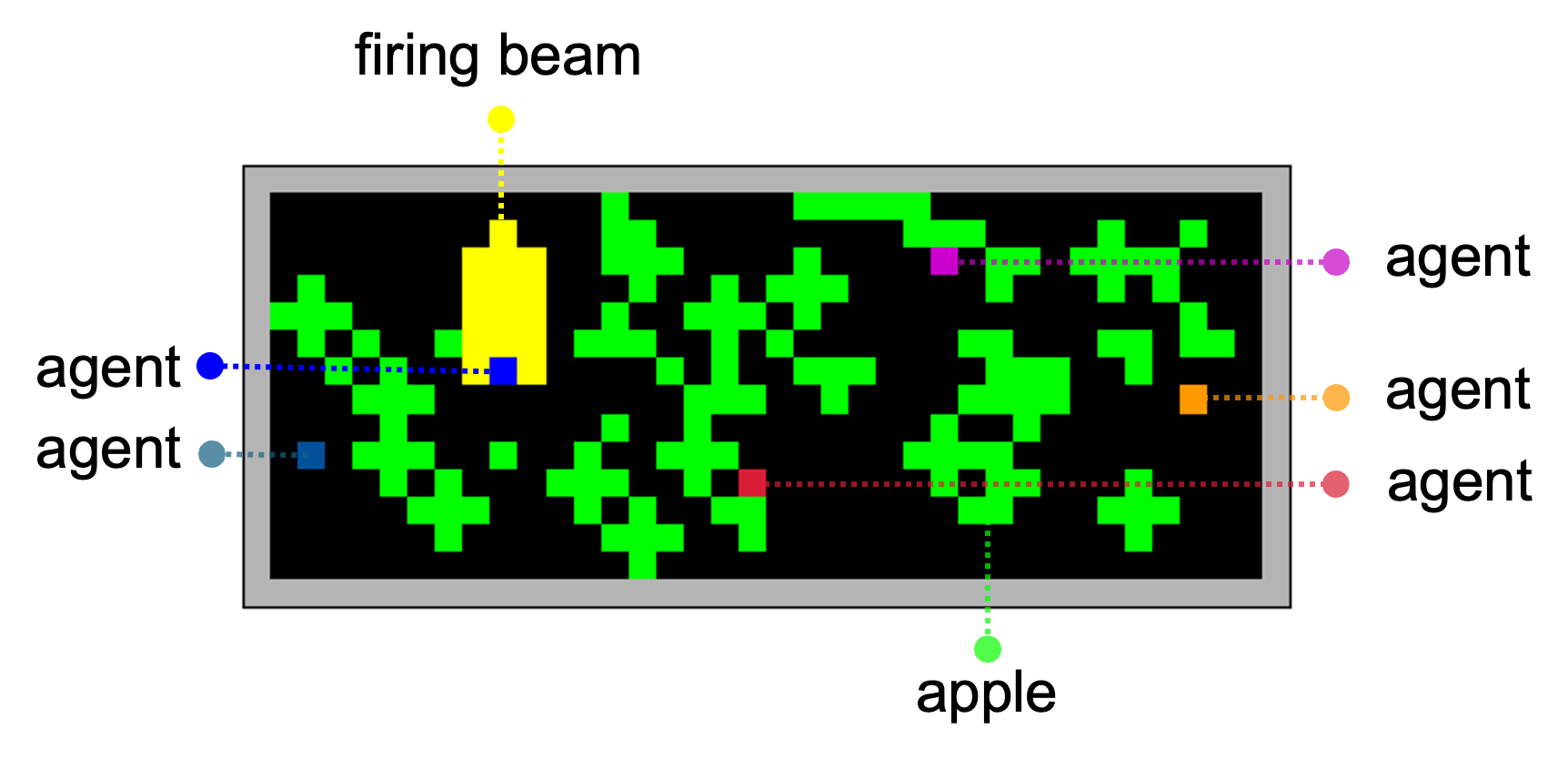}
    \caption{Illustration of Harvest \citep{hughes2018inequity}.}
    \label{fig:harvest}
\end{figure}

\paragraph{Harvest \textup{\citep{hughes2018inequity}}} 
collecting apple provides a reward of 1. 
The apple regrowth rate varies across the map, dependent on the spatial conﬁguration of uncollected apples: the more nearby apples, the higher the local regrowth rate. 
If all apples in a local area are harvested then none ever grow back until the game is reinitialized. 
Agents can fire penalty beams, inflicting a reward of -50 reward on the hit agent at the cost of a -1 reward to themselves.
The illustration of Harvest is shown in \Cref{fig:harvest}.

\section{Multi-Agent Cooperation in SSDs}

While multi-agent cooperation in sequential decision-making has been a long-standing topic in MARL that takes stochastic games as the standard model, this survey features a narrowed-down focus on multi-agent cooperation in SSDs. For multi-agent cooperation in stochastic games using MARL, we refer readers to 
\citep{busoniu2008comprehensive} for a comprehensive survey.

\citet{leibo2017multi} pioneered SSDs, defining the framework and proposing a modeling approach connecting two-agent SSDs and normal-form SDs through empirical game-theoretic analysis (EGTA). 
EGTA is a technique that employs agent-based simulation to generate data from game environments, thereby modeling game that are difficult to analyze \cite{wellman2024empirical}.
In this work \citep{leibo2017multi}, cooperative and defective policies are trained using independent deep reinforcement learning under various environmental settings. 
These policies are then evaluated to construct empirical payoff matrices,
which exhibit payoff relationships similar to that of normal-form SDs.
However, from a learning perspective, SSDs differ from normal-form SDs as that agents in SSDs must 
learn effective execution of cooperation or defection rather than just learn to select between them.

The challenges in solving SSDs include developing usable policies in stochastic games with large state and policy spaces as mentioned above, as well as addressing sub-optimal equilibrium that independently maximizing the individual interest of each agent may lead to counterproductive outcomes in SSD. 
Fortunately, with the powerful generalization and representational ability of deep neural networks, MARL offers a promising solution to develop usable policies in high-dimensional stochastic games \citep{hernandez2019survey}.
Therefore, we focus on addressing sub-optimal equilibrium in SSDs, which we call \textit{solve SSDs}.
This section reviews prevailing MARL-based works on solving SSDs on two lines as shown in \Cref{fig:SSD problem} (B). 
One line of research takes the planner's view, which is concerned with how to promote multi-agent cooperation for higher social welfare by designing suitable reward functions through shaping intrinsic and external motivations.
The other line adopts the participant's view and tries to design agents to play against opponents with uncertain diverse policies, in order to exploit them or avoid being exploited by them.
Their key difference lies in their perspectives on solving SSDs.

\subsection{Shaping Intrinsic and External Motivations}\label{sec:31}

In SSDs, achieving cooperation leads to higher collective long-term returns for all agents than each agent acting selfishly. 
In order to encourage agents to learn cooperative policies, it is usually necessary to introduce additional motivation rewards in addition to environmental rewards.
Most works in this line introduce the motivation by shaping agent $i$'s reward $r_i^{\text{total}}$ as the weighted sum of the environmental reward $r_i^{\text{env}}$ and the motivation reward $r_i^{\text{mot}}$ like the following form:
\begin{equation}
    r_i^{\text{total}}=\alpha r_i^{\text{env}}+\beta r_i^{\text{mot}},\nonumber
\end{equation}
where $\alpha$ and $\beta$ are weights.
According to the sources of the motivation reward, we now review the mechanisms for shaping motivations in SSDs from two perspectives: the \textit{intrinsic motivation} and the \textit{external motivation}.
\Cref{tab:summary1} summarizes the works cited in \Cref{sec:31}.

It is worth noting that the aim of works on mechanisms is to validate whether the introduced mechanisms can enhance cooperation in SSDs, typically by comparing the level of cooperation with and without these mechanisms. 
Only a few works compare the performance across different mechanisms \citep{li2023learning,hua2023learning}. 
This is due to the varying assumptions and information requirements of each mechanism, which make fair comparisons challenging.


\subsubsection{Intrinsic Motivation}
We refer to motivations that agents not only care about one's own but also the others' reward as the intrinsic motivation. 
The idea of the intrinsic motivation in SSDs is rooted in behavioral economics and social psychology, in which research has found that individuals have the intrinsic motivation to do certain things for their inherent satisfactions rather than for some specific consequence \citep{ryan2000intrinsic}, e.g., inequity aversion, social value orientation, altruism, social influence, and reputation.

\begin{table*}[ht]
\centering
\caption{Summary of works in shaping intrinsic and external motivations in SSDs.}
\label{tab:summary1}
\begin{tabular}{@{}lll@{}} 
\toprule
\textbf{Motivation} & \textbf{Mechanism} & \textbf{Work} \\ 
\midrule
Intrinsic Motivation & Inequity Aversion & Weighted inequity aversion [\citet{hughes2018inequity}] \\
& Altruism & Evolving shared reward network [\citet{wang2019evolving}]\\
&          & Team credo [\citet{radke2023importance}]\\
&          & Organizational psychology[\citet{radke2022exploring}] \\
&          & Team size [\citet{radke2023towards}]\\
& Social Value Orientation & Driver behavior estimating with SVO[\citet{schwarting2019social}]\\
&                          & SVO and cooperation [\citet{mckee2020social}]\\
&                          & SVO and generalization[\citet{madhushani2023heterogeneous}] \\
&                          & RESVO [\citet{li2023learning}]\\
&                          & RUSP [\citet{baker2020emergent}]\\
& Social Influence & Causal inﬂuence [\citet{jaques2019social}]\\
& Reputation & Reputation of contribution [\citet{mckee2021deep}]\\
\hline
External Motivation& Peer Rewarding & Gifting [\citet{lupu2020gifting}]\\
&& Share trading [\citet{schmid2022learning}]\\
&& LIO [\citet{yang2020learning}]\\
&& Taxation [\citet{hua2023learning}]\\
& Agreement & Social norm [\citet{vinitsky2023learning}]\\
&& Contract [\citet{vinitsky2023learning}]\\
&& Price of anarchy [\citet{gemp2022d3c}]\\
&& Sociality matching [\citet{eccles2019learning}]\\

\bottomrule
\end{tabular}
\end{table*}

\paragraph{Inequity aversion}
\citet{hughes2018inequity} pioneer the incorporation of intrinsic motivation from behavioral economics and psychology explicitly within SSDs, and define it through the lens of inequality aversion. 
Inequity aversion is the preference for fairness and resistance to incidental inequalities \citep{fehr1999theory}, which consist of advantageous inequity aversion and disadvantageous inequity aversion.
To provide a glimpse of the definition of intrinsic motivation in SSDs, we present Hughes's formalization based on inequality aversion as follows:  
\begin{equation*}
    \begin{aligned}
    r_i^{\text{mot}}=& -\frac{\alpha^i}{N-1}\sum\nolimits_{j\ne i}\max\{e^j_t(o^j_t,a^j_t) 
    -e^i_t(o^i_t,a^i_t),0\}\\
    &- \frac{\beta^i}{N-1}\sum\nolimits_{j\ne i}\max\{e^i_t(o^i_t,a^i_t)-e^j_t(o^j_t,a^j_t),0\},
\end{aligned}
\end{equation*}
where $o^i_t$ is agent $i$'s  observation of the global environmental
state $s_t$, and the agent $i$'s temporal smoothed rewards $e^i_t$  are updated by $e^i_t(o^i_t,a^i_t)=\gamma \lambda e^i_{t-1}(o^i_t,a^i_t)+r_i^{\text{env}}(o^i_t,a^i_t)$.
Intuitively, the first term characterizes the agent $i$'s reward loss when other agents achieve rewards greater than agent $i$'s own and thus represents the disadvantage inequity aversion, and the second term characterizes the reward loss when agent $i$ performing better than others and thus represents the advantage inequity aversion.
 $\alpha^i$ and $\beta^i$ control agent $i$'s aversion to disadvantageous inequity and advantageous inequity.
Interestingly, the authors find that advantageous inequity aversion promotes cooperation in public goods dilemmas by providing an unambiguous feedback signal, while disadvantageous inequity aversion promotes cooperation in common dilemmas by providing a noisy signal for agents' cooperative and sustainable behaviors.

\paragraph{Altruism}
Altruism amounts to taking costly actions that mostly benefit other individuals \citep{kerr2004altruism}.
In SSDs, altruistic agents prioritize other agent's reward or the collective reward of the entire group over their own individual rewards.
\citet{wang2019evolving} introduce the altruistic intrinsic motivation parameterized by a shared reward network, which is updated slowly compared to policy networks to simulate natural selection for a higher collective reward of all agents. 
Without handcrafting reward shaping, their altruistic evolutionary paradigm solves difficult SSDs.
Some works assume that individual altruism only applies to certain members within a group, which divides the entire group into different teams to promote cooperation more effectively.
\citep{radke2023importance,radke2022exploring}.
\citet{radke2023towards} discuss advantages and disadvantages of team division on full-group cooperation.
They find the optimal size for team division to promote cooperation in the entire group and prove it both theoretically and experimentally.

\paragraph{Social value orientation}
Social Value Orientation (SVO) measures how an agent apportion rewards of itself and others in its reward function\citep{mcclintock1989social,murphy2011measuring}.
Aforementioned inequity aversion and altruism can be consider as particular cases of SVO.
Here we focus on works that study impact of general SVO rather than considering only specific orientation.
\citet{schwarting2019social} highlight the nature of SDs in autonomous driving and enhance the prediction of other drivers' behaviors by estimating their SVOs.
\citet{mckee2020social} first use SVO to characterize the social preferences of agents in SSDs. 
The relationship between the agent $i$'s own reward $r_i$ and the average reward of other agents $\overline{r}_{-i}$ is characterized by a reward angle $\theta (\vec{R})=\arctan \left(\overline{r}_{-i}/r_i\right)$.
By defining the motivation reward  $r_i^{\text{mot}}$ as $-\omega\cdot \lvert \theta^{\text{SVO}}-\theta(\vec{R}) \rvert$, the agents with specific SVO given by $\theta^{\text{SVO}}$ can be trained.
In addition, they also find that populations trained with heterogeneous SVO develop more universal policies with higher levels of cooperation than homogeneous populations.
Furthermore, \citet{madhushani2023heterogeneous} find that learning best responses to diverse policies trained with heterogeneous SVO leads to better zero-shot generalization in SSDs.
\citet{li2023learning} propose a learning framework based on SVO called RESVO, which emerges stable roles in populations and efficiently solves SSDs through the division of labor. 
In addition, similar to SVO,  \citet{baker2020emergent} propose an environment augmentation called Randomized Uncertain Social Preferences (RUSP)  to characterize social preferences,
and find emergent direct reciprocity, indirect reciprocity and reputation, and team formation when training agents with RUSP.

\paragraph{Social influence}
Many experimental studies have revealed that individual behavior is subject to social influences \citep{cialdini2004social,asch2016effects}.
\citet{jaques2019social} introduce social inﬂuencers into SSDs, who use social influence as their intrinsic reward.
The social influence is defined as a causal inﬂuence: to what extent the other agents change their actions because of the influencer's action.
This is similar to the concept of informational social influence, which is defined as an influence to accept information obtained from another as evidence about the nature of the game \citep{deutsch1955study}.
High level of cooperation in SSDs within three social influence models encompassing basic inﬂuence model, inﬂuential communication and modeling other agents demonstrates that introducing of social influencers leads to enhanced coordination and communication among agents.
\paragraph{Reputation}
Reputation serves as a crucial reciprocal mechanism for fostering cooperation within large-scale human populations \citep{qalati2021effects}.
\citet{mckee2021deep} model one's reputation based on the matching level of its own and group contributions in SSDs.
The authors conduct MARL simulations and human behavior experiments under identifiable (contributions of each group member can be perfectly monitored) and anonymous (contributions can not be perfectly monitored) conditions.
Results of several metrics of agents' behavior demonstrate that reputation promotes cooperation among agents by developing a non-territorial, turn-taking strategy to coordinate collective action.

\subsubsection{External Motivation}
In contrast to intrinsic motivation, external motivation is a construct that pertains whenever an activity is done in order to attain some separable outcome \citep{chentanez2004intrinsically}.
In this survey, we refer to motivation directly from other agents as external motivation.

\paragraph{Peer rewarding}
Peer rewarding is a type of external motivation, allowing agents' reward functions to be directly modified by others. 
Technically, peer rewarding improves reward sparsity and encourages agents to explore cooperative behaviors in SSDs.
\citet{lupu2020gifting} introduce gifting options into agents' action space and find that zero-sum gifting solves the tragedy of the commons most effectively. 
They introduce three types of gifting: zero-sum, where agents can send a gift $g$ but incur an immediate penalty of $-g$ from its gifting; fixed budget, where agent can gift $g$ from a fixed budget $B$ in each episode; and replenishable budget, where gift budget can increment as the quantity of rewards it can collect from the environment.
Further, some work study learnable peer rewarding.
\citet{yang2020learning} propose the Learning to Incentivize Others (LIO) framework, where each agent learns its own incentive function by explicitly accounting for its impact on recipients’ behavior directly and its own extrinsic objective indirectly.
They find that LIO agents ﬁnd a near-optimal division of labor in SSDs.
In addition, some economic concepts, such as market participation \citep{schmid2022learning}, where agents learn to participate in others' rewards by acquiring shares, and taxation \citep{hua2023learning}, where a centralized agent learns the Pigovian tax/allowance to maximizes the social welfare, 
have also been introduced into SSDs to model peer rewarding.


\paragraph{Agreement}
Human societal structures are under-girded by a multitude of agreements, which are designed to incentivize compliance through rewards for those who conform to their mandates, and to enforce discipline through penalties for those who violate the same.
Can these agreements also promote cooperation among AI agents in SSDs? 
\citet{vinitsky2023learning} construct an agent architecture called Classiﬁer Norm Model (CNM) that can use public sanctions to spark the emergence of social norms in SSDs.
The CNM agents learn to classify others' transgression and enforce social norms from experience, and they converge on beneficial equilibrium and are better at resolving free-rider problems.
\citet{christoffersen2023get} consider the contracting augmentation among agents, where agents voluntarily agree to binding state-dependent transfers of reward.
\citet{gemp2022d3c} construct D3C agents to minimize the price of anarchy, a gap between the welfare that can be achieved through perfect coordination against that achieved by self-interested agents at Nash equilibrium. 
D3C resembles the celebrated Win-Stay-Lose-Shift \citep{nowak1993strategy} strategy to improve inefficient equilibrium in SSDs.
\citet{eccles2019learning} construct online-learning reciprocal agents, who try to measure and match the level of sociality of others and influence naive agents to promote their cooperation. 
The results indicate that reciprocal agents can influence naive agents to promote their cooperation.

\begin{table*}[ht]
\centering
\caption{Summary of works in playing against diverse opponents in SSDs.}
\label{tab:summary2}
\begin{tabular}{@{}lll@{}} 
\toprule
\textbf{Category} & \textbf{Work} \\ 
\midrule
Adaptive Policy Adopting  & amTFT [\citet{lerer2017maintaining}] \\
& Cooperation degree detection [\citet{wang2019achieving}]\\
& Graph-based TFT [\citet{gleau2022tackling}]\\
& Tunable agent [\citet{o2021exploring}]\\
\hline
Opponent Shaping & LOLA [\citet{foerster2018learning}]\\
& SOS [\citet{letcher2018stable}]\\
& COLA [\citet{willi2022cola}] \\
 & POLA [\citet{zhao2022proximal}]\\
& M-FOS [\citet{lu2022model}]\\
& Meta-Value Learning [\citet{cooijmans2023meta}]\\
\bottomrule
\end{tabular}
\end{table*}

\subsection{Playing against Diverse Opponents}\label{sec:32}
The other paradigm of solutions to SSDs is playing with diverse opponents. 
One may feel this resembles solving a zero-sum game at the first glance, but the difference is that in mixed-motivation SSDs, directly maximizing one's own rewards may have opposite effects. 
In fact, it is more reasonable to decide one's own choice based on the opponent's behavior style, as is verified in the famous strategies Tit-for-Tat (TFT) or Win-Stay-Lose-Shift (WSLS) in iterated SDs.
Based on the principles of constructing strategies, we divide the prevailing work into two categories: two-stage adaptive policy adopting and online learning opponent shaping. 
\Cref{tab:summary2} summarizes the works cited in \Cref{sec:32}.
\paragraph{Adaptive policy adopting} 
This type of method typically involves two stages: first pre-training different styles of policies; then adaptively selecting by designed strategies in real-time games.
\citet{lerer2017maintaining} construct an approximate Markov Tit-for-Tat (amTFT) agent.
They train cooperative and defective policies by reward shaping, and design a TFT-like strategy manually to decide when to cooperate or defect. 
Further, \citet{wang2019achieving} propose the cooperation degree detection network, which is a classifier trained on trajectories generated by pre-trained policies with various degrees of cooperation. 
Experimental results show that the strategy can avoid being exploited by exploitative opponents and achieve cooperation with cooperative opponents.
\citet{gleau2022tackling} introduce Circular SSD and construct Graph-based TFT for asymmetric game scenarios.
Moreover, \citet{o2021exploring} propose a method of training agents with tunable levels of cooperation in SSDs, based on multi-objective reinforcement learning.

\paragraph{Opponent shaping}


An opponent shaping agent tries to influence the future strategies of its learning opponents online by offering feedback on their behavior.
\citet{foerster2018learning} propose the Learning with Opponent Learning Awareness (LOLA) rule, which takes into account the learning dynamics of the opponent.
In detail, the LOLA learning rule includes an additional term that accounts for the impact of one agent’s policy on the anticipated parameter update of its opponent:
\begin{equation}
    V^1(\theta^1,\theta^2+\Delta \theta^2)\approx V^1(\theta^1,\theta^2)+(\Delta \theta^2)^T \nabla_{\theta^2}V^1(\theta^1,\theta^2),\nonumber
\end{equation}
where $\nabla$ is the gradient operator, $V^i(s;\theta^1,\theta^2)$ is the state-value function (expected cumulative rewards with the initial state $s$) of agent $i$ parameterized by two agents' policy parameters $\theta^1$ and $\theta^2$, and the update of $\theta^2$ follows $\Delta \theta^2=\eta \nabla_{\theta^2}V^2(s;\theta^1,\theta^2)$ with $\eta >0$ being the step size.
The authors demonstrate that LOLA can achieve reciprocity-based cooperation under self-play settings by finding TFT-like strategies and also exploit other naive learning algorithms. 
Subsequently, a series of improvements to LOLA are proposed.
\citet{letcher2018stable} propose Stable Opponent Shaping that interpolates robustly between LOLA and LookAhead \citep{zhang2010multi} to prevent LOLA agents from arrogant behavior and converging to non-ﬁxed points.
\citet{willi2022cola} propose Consistent LOLA to address the consistency problem of LOLA.
And \citet{zhao2022proximal} introduce Proximal LOLA which guarantees behaviorally equivalent policies result in behaviorally equivalent updates to address LOLA's sensitivity to policy parameterization.
Furthermore, in order to address the limitation of explicit gradients required by the LOLA series algorithms, \citet{lu2022model} propose Model-Free Opponent Shaping (M-FOS) that learns in a meta-game in which each meta-step is an episode of the underlying (``inner") game. The meta-state consists of the inner policies, and the meta-policy produces a new inner policy to be used in the next episode. 
In addition to policy-based method, a value-based opponent shaping method called Meta-Value Learning is proposed by \citet{cooijmans2023meta}, which can be seen as the value-based complement to the policy gradient of M-FOS.

\begin{table*}[ht]
\centering
\caption{Summary of works in human-agent cooperation and leveraging agents to motivate human-human cooperation in SDs.}
\label{tab:summary3}
\begin{tabular}{@{}lll@{}} 
\toprule
\textbf{Subject} & \textbf{Motivation} & \textbf{Work} \\ 
\midrule
Human-agent  & Algorithms design & S++ and S\# [\citet{crandall2014towards, crandall2018cooperating, oudah2018ai}] \\
cooperation& Exploitation avoiding  & Zero-determinant strategy [\citet{hilbe2014extortion, wang2016extortion}]\\
&          & Safety control [\citet{zhangRethinkingSafeControl2023}]\\
& Experiment-driven insights   & Partner selection [\citet{santosOutcomebasedPartnerSelection2019, santosPickyLosersCarefree2020}] \\
& Human bias towards agents & Transparency-efficiency tradeoff [\citet{ishowo2019behavioural}]\\
&                          & Algorithm exploitation [\citet{karpus2021algorithm}]\\
&                          & Guilt and envy [\citet{melo2016people}] \\
& Human bias mitigating     & Culture and emotion [\citet{de2019cooperation}]\\
&                          & Verbal communication [\citet{maggioni2023if}]\\
\hline
Human-human & Network engineering & Social planner [\citet{shirado2020network, mckee2023scaffolding}]\\
cooperation & Evolutinary dynamics  & Fixed behavior [\citet{terrucha2022art, guoFacilitatingCooperationHumanagent2023, sharma2023small}]\\
&                          & Opportunity cost [\citet{hanWhenNotTrust2021}]\\
&Authority grant & Considering long-term profit [\citet{de2019human}]\\
&                & Predefined and customized agents [\citet{fernandez2022delegation}]\\

\bottomrule
\end{tabular}
\end{table*}


\section{Human-Agent Cooperation in SDs}\label{sec:4}

In the era of artificial intelligence, technologies like autonomous vehicles \citep{faisal2019understanding}, chatbots \citep{lin2020caire}, and human-computer interaction systems \citep{preece1994human} have become ubiquitous. AI agents now handle delicate and complex tasks, ranging from disaster relief to city construction. While extensive research has explored how agents coordinate with humans - a field known as human-agent coordination \citep{nikolaidis2015efficient, chugunova2022we, bonnefon2024moral} - there are also scenarios where conflicts of interest emerge between humans and agents. Human-agent coordination focuses on shared interests, where agents' goals align perfectly with those of humans. In contrast, human-agent cooperation centers on exploring strategies to sustain high cooperation levels in human-agent hybrid systems amidst strategic conflicts and SDs. Despite overlaps between coordination and cooperation with agents, the primary challenges diverge. This section explores human-agent cooperation in SDs from two angles: (i) developing algorithms that effectively enhance cooperative behavior; (ii) the biases humans exhibit in their decision-making processes when interacting with both humans and agents.
We summarize the references used in \Cref{sec:4} at the top of \Cref{tab:summary3}.

\subsection{Designing Algorithms for Cooperation with Humans}

\paragraph{From S++ to S\#} 


Successful AI agents require three key properties: generality across various games, adaptability to unfamiliar counterparts, and swift learning in decision-making \citep{crandall2018cooperating}. The S++ and S\# family addresses these needs, focusing on iterated interactions where individuals repeatedly engage in the same game structure and make decisions based on past experiences. \citep{crandall2014towards, oudah2018ai}. 


S++ \citep{crandall2014towards}, recognized as an expert algorithm, is originally designed for iterated normal-form games. It stands out with two main components: firstly, its set of experts, which are systematically generated from the game's description, and secondly, its expert-selection mechanism, which dictates the choice of expert to be utilized in each round of the game. Building upon the S++ algorithm, \citet{crandall2018cooperating} systematically investigate human-agent cooperation in iterated normal-form games (e.g., PD and Chicken games) and stochastic games (e.g., Stochastic Game Prisoner’s Dilemma and Block Game). 
In the interaction with other algorithms such as generous Tit-for-Tat, Win-Stay-Lose-Shift, etc, the S++ algorithm emerges as the top performer. A human-agent experiment subsequently reveals that S++ is as proficient at establishing cooperative relationships with humans as humans are. However, S++ and humans often fail to achieve high cooperation levels with human partners. Therefore, a communication mechanism is introduced. 

The S\# algorithm \citep{crandall2018cooperating}, an advancement of the S++ algorithm, integrates a communication framework that allows agents to engage in and respond to cheap talk. Its capacity to enhance cooperation has been validated through both simulations and human-agent experiments. The results showcase its ability to cooperate with humans and other algorithms at levels that rival human cooperation. 
Subsequently, \citet{oudah2018ai} further develops the S++ and S\# algorithms by introducing diverse signaling strategies inspired by various philosophical approaches. This amalgamation of behavior strategy (the agent's actions) and signaling strategy (the agent's verbal communication) endeavors to optimize player utility by influencing people and winning friends. Findings reveal that Carnegie’s Principles \citep{carnegie2023win}, a signaling strategy focused on refraining from criticism, complaints, or condemnation while positively uplifting partners, coupled with a behavior strategy characterized by rapid and effective learning, proves most adept at winning friends and influencing people.

\paragraph{Avoid exploitation by self-seeking individuals.} 
Mitigating the obstacle of cooperation requires handling human selfishness. Zero-determinant (ZD) strategy becomes a viable option. ZD strategies establish a linear payoff correlation between a focal player and the counterpart, irrespective of the counterpart's strategy \citep{press2012iterated}. This category includes extortion strategies, ensuring that a player's surplus consistently surpasses the counterpart's surplus by a fixed percentage, and generosity strategies, designed to incentivize cooperation from the counterpart. Building upon extortion and generosity strategies, \citet{hilbe2014extortion} conduct a human-agent experiment and observe that although extortion strategies do prevail over their human counterparts, this success is met with retaliation, leading humans to choose defection in response. Consequently, generosity emerges as the more profitable and effective strategy. In contrast, extortion strategies outperform generosity strategies when there is a long enough duration for interaction and when human participants are aware they are competing against a computerized counterpart \citep{wang2016extortion}. 
Human-agent cooperation finds application in safety-critical interactions between humans and agents. \citet{zhangRethinkingSafeControl2023} find that conventional safe control methods may not consistently achieve the highest safety and performance when faced with self-seeking humans. Addressing the issue of myopic behaviors, they propose an algorithm incorporating the influence of risk and long-term reward. The algorithm can prevent autonomous agents from being exploited by self-seeking humans and seeks to achieve a more optimal balance between safety and performance in human-agent interactions.

%

 
\paragraph{Experiment-driven insights} Designing agents that promote cooperation can be inspired by experimental findings.
\citet{santosOutcomebasedPartnerSelection2019} conduct human-agent experiments involving human participants who interact with two programmed agents in a collective risk dilemma. Following the game, participants complete questionnaires regarding their preferences for future robotic partners. The results from these post-task questionnaires indicate a tendency among humans to favor cooperating agents following collective failures. Conversely, there is no distinct preference following collective successes. This observation prompts the development of an outcome-based partner selection model, further explored through an evolutionary game theory framework to evaluate its effectiveness. An online human-agent study reinforces these findings and highlights the significant influence of outcome-based strategies in human-agent interactions \citep{santosPickyLosersCarefree2020}.



\subsection{Revealing Human Bias towards Agents}



\paragraph{Bias caused by the nature of counterparts} 

Within human-agent systems, evidence has shown that the revelation of an agent's true identity holds sway over human behavior \citep{ishowo2019behavioural, karpus2021algorithm}. \citet{ishowo2019behavioural} conduct an intriguing experiment involving iterated PD games, wherein participants or agents engage in continuous interactions with a designated counterpart. The crux of their investigation lies in probing the delicate tradeoff between transparency-revealing one's identity as an agent—and efficacy-the capacity to cooperate seamlessly in human-agent interactions. Human participants, crucially, are divided into groups: one accurately briefed on the nature of their counterparts—whether human or agent and the other misled by false information.
Employing the S++ algorithm, agents are more successful in eliciting human cooperation when their non-human identity is undisclosed. Yet, this advantage diminishes once their non-human status is revealed. The bias against agents persists and cannot be mitigated over time, even after humans realize that agents are more efficient in promoting human cooperation. 
Moreover, experiments involving SD games with human players and agent counterparts have shown that even when humans expect cooperation from their counterparts, they are more likely to exploit an agent's benevolence than that of another human \citep{karpus2021algorithm}. Notably, the agent's behavior is engineered to mirror human actions in these games. Empirical findings reveal a noteworthy trend: participants don't demonstrate diminished trust in agents relative to humans. Instead, they display a heightened tendency to exploit agents, particularly when anticipating cooperation from the opposing party.
Such distinctions are also prevalent in emotional responses, particularly those related to feelings of guilt and envy towards agents and humans. \citet{melo2016people} reveal that humans exhibit similar levels of envy when interacting with both agents and humans, but 
report significantly less guilt in interactions with agents within SD games.

\paragraph{Mitigating human bias} Cultural cues, emotion, and verbal communication hold promise for mitigating human biases in interactions with agents. \citet{de2019cooperation} organize human-agent experiments wherein humans participate in iterated PD with agents featuring virtual faces and emotional expressions. The findings unveiled a compelling insight: cultural and emotional attributes have the potential to diminish biases, fostering more effective and cooperative interactions. Subsequently, \citet{maggioni2023if} delve deeper into the realm of human-agent interaction within iterated PD games, demonstrating the influential role of verbal communication in fostering cooperative strategies and partially mitigating biases against agent counterparts. Together, these revelations offer a glimmer of hope in mitigating biases against agent counterparts.

\section{Inspirations on Human-Human Cooperation}


In human-agent systems, agents scaffold human-human cooperation by assuming various roles. These include acting as planners to structure network interactions, operating as independent decision-makers to affect population composition, and making decisions on behalf of humans. We summarize the references used in Section 5 at the bottom of Table 3.


\paragraph{Engineering network structure} Social networks play a key role in overcoming SDs. \citet{shirado2020network} utilize intervention AI agents to reshape local social connections between humans. The results reveal a significant increase in human-human cooperation when agents employing disengaged intervention (cutting a defective neighbor) are introduced. Moreover, even a single agent implementing a mixed strategy (including engaged, disengaged, and self-rewiring intervention) significantly alters social dynamics and improves cooperation, fostering the development of cooperative clusters. Subsequently, \citet{mckee2023scaffolding} explore human-human cooperation using deep reinforcement learning in network games. They train an agent as a GraphNet planner, responsible for guiding humans in creating or breaking interaction links. This planner consistently recommends building links between cooperators and discourages new links with defectors. Notably, the planner adopts a conciliatory strategy to address defectors, initially establishing a certain number of cooperate–defect links and progressively suggesting the severance of these links over time.

\paragraph{Influencing evolutionary dynamics}

In human-agent populations, the evolution of cooperation hinges significantly on the behavior, proportion, and spatial distribution of the agents. \citet{terrucha2022art} utilize evolutionary game theory to model the interactions and strategies of adaptive (human-like) and fixed-behavior agents in collective risk dilemma games. Their findings suggest that the presence and behavior of agents significantly affect human cooperation levels, with humans adjusting their strategies to compensate for the agents' actions. Motivated by this, \citet{guoFacilitatingCooperationHumanagent2023} investigate the impact of cooperative and defective agents on human cooperation. Employing replicator dynamics and pairwise comparison, the study shows cooperative agents have varying impacts, with limited influence in prisoner's dilemma games but promoting cooperation in stag hunt games. Defective agents, intriguingly, can lead to complete dominance of cooperation in snowdrift games. Additionally, the role of population structure and imitation strength is highlighted as crucial in these dynamics. Agents with a purely constant strategy can also efficiently encourage human cooperation. \citet{sharma2023small} discover that certain types of agents, such as always-cooperate agents and loner agents, significantly promote cooperation under different structural conditions. In particular, this study identifies an optimal density of loner agents for maximizing cooperation, suggesting that the diversity in agent actions can diminish their cooperation-promoting effect. On the other hand, in situations where humans lack transparency regarding the actions of agents, \citet{hanWhenNotTrust2021} underscore the importance of considering the opportunity cost when adopting a reciprocal strategy. In response to this concern, they develop a trust-based strategy that, upon establishing trust, utilizes probabilistic checks to minimize potential opportunity costs, demonstrating its effectiveness in maintaining cooperation.


\paragraph{Making decisions on behalf of humans} Humans have the capacity to entrust decision-making authority to agents. Particularly, this mode of interaction can change the way humans solve these SDs. \citet{de2019human} find that humans exhibit higher cooperation when decisions are agent-driven rather than made personally in games with repeated interactions. The rationale is that programming encourages individuals to consider long-term interests while diminishing the influence of immediate short-term self-interest, thereby fostering greater cooperative behavior. 
\citet{fernandez2022delegation} also highlight the positive impact of delegating decision-making to AI agents in collective risk scenarios, including two forms of delegation: to AI agents with predefined behaviors and to customizable AI agents whose behaviors are defined by participants. Both forms significantly enhanced cooperation and group success in achieving collective targets, compared to groups composed solely of humans. Notably, customizable AI agents led participants to adopt cooperative strategies more frequently, underscoring the potential of AI as a tool for promoting prosocial behavior.







\section{Directions of Future Research}
While fruitful efforts have been put into the investigation of cooperation in SDs, several aspects have yet to be explored in the domain, which are summarized below.

\smallskip

{\bf (1)} {\em Enhancing cooperation in multi-agent and human-agent settings with large language models (LLMs)}.
LLMs are generative AI models trained on extensive human textual data \citep{zhao2023survey}.
LLMs excel in processing and generating natural language text, which makes them useful for a wide range of language tasks.
Therefore, LLMs have been employed to construct large social simulation systems with higher interpretability \citep{park2023generative,ren2024emergence}.
When focusing on the study of cooperation in SDs, LLMs might empower the study in the following two ways.
Firstly, they can serve as a source of prior knowledge for shaping agent rewards.
LLMs have been employed to generate rewards or reward functions for RL agents in recent works, steering RL agents to coordinate with humans better or achieve human-level precision control \citep{kwon2022reward,hu2023language,yu2023language,ma2023eureka}.
Nevertheless, it remains an open question how RL agents that learn from LLM-shaped rewards to solve SDs effectively.
The second usage is to employ LLMs as the backbone of agent architectures for generative agents in SDs. 
Some recent works have evaluated the traits and behavioral characteristics of LLMs in SD games (e.g. PD, SH and HD), but their conclusions reached are not always consistent \citep{akata2023playing, brookins2023playing, lore2023strategic, fan2024can,mei2024turing}.
The differences in their conclusions may be due to variations in the prompts they used.
Continuing to explore the capabilities and performance of LLMs in SDs is crucial, especially as we tackle additional challenges in these issues like building reputation, identifying deception, and avoiding exploitation, as well as encouraging cooperation through potential punitive measures.

\smallskip

{\bf (2)} {\em Establishing a theoretically guaranteed framework for fostering and sustaining cooperation in SDs.} While methods surveyed in this paper show empirical promise in encouraging cooperation in SDs, none offers theoretical guarantees. A theoretical framework is envisioned to bridge past and future SD studies: precisely explaining existing empirical findings and guiding the design of more effective methods. It is crucial to note that conditions for maintaining cooperation differ between normal-form SDs and SSDs \citep{kleshnina2023effect}, necessitating a separate consideration in constructing the theoretical framework. Moreover, for human-agent hybrid systems, existing studies are mostly empirical those exceptions that leverage evolutionary game theory cannot capture human biases towards AI agents, a critical aspect of human-agent hybrid systems, and thus fail to provide an accurate prediction or explanation of the evolution of cooperation in these systems.
It is worth emphasizing that when handling behavioral experiment data from human users, preserving user privacy through theoretical methods is also crucial\citep{liu2020differentially,gao2024private,gao2024differentially}.

\smallskip

{\bf (3)} {\em Further employing SDs to model real-world decision-making scenarios.} Large-scale simulation systems on economics, sociology and ecology with diverse agents, e.g., the board game Diplomacy \citep{kramar2022negotiation}, offer insights into the mechanism of system evolution. These low-cost, realistic simulations better inform policymakers and even inspire solutions to relief SDs. As AI agents are integrated with humans, they provide convenience whilst introduce possibly negative social influence to this human-agent hybrid system. A typical real-world scenario is autonomous driving, where AI agents face dilemmas like choosing between speed and safety at intersections and must navigate undesirable selfish behavior from human drivers. 
Current SD research often simplifies driving scenarios \citep{schwarting2019social}. As autonomous vehicles being integrated with humans, infusing them with realistic social characteristics is crucial for positive societal impact. Addressing collision risks among autonomous and human-operated vehicles necessitates designing algorithms for large-scale human-agent hybrid systems and constructing cost-effective, extensive simulation systems.
More broadly, SDs holds promise to build simulation systems spanning economics, sociology, or ecology can simulate AI agents with diverse social preferences, offering insights into system evolution. These realistic simulations can better inform policymakers. However, it is important to note that interpretations of AI simulation outcomes requires a dialectical perspective, and vigilance toward ethical concerns is paramount.
All in all, such SDs-based AI agents must adeptly respond to human behaviors and biases, underlining the importance of developing human-agent hybrid systems for realistic scenarios to enhance human-agent cooperation and tackle key cooperative challenges in SDs.
\smallskip

{\bf (4)} {\em Bridging human-agent cooperation and SSDs.} Compared to the fruitful literature on multi-agent cooperation in SSDs, the investigation of human-agent in SSDs is insufficient as of this survey, though with only few exceptions \citep{crandall2018cooperating}. In SSDs, the process of decision-making is extended over a series of steps, consequently leading to human behavioral patterns that often diverge from those observed in the single-stage settings (e.g., normal-form SDs) \citep{sims2013melioration}.  
This brings new challenges to human-agent cooperation: It will become difficult to develop AI systems that can nudge humans towards cooperation by avoiding being myopic and focusing on longer-term benefits, and to understand the behavioral tendencies humans exhibit in sequential decision-making processes. 
In SSD scenarios, people must consider both their behavior rules, the behavior of agents, and their interactions with the environment, which form a complex feedback loop. For example, cooperative behaviors can cultivate a peaceful or replete environment, leading to long-term benefits. However, the need to maximize immediate profits adds complexity to decision-making. This complexity is further driven by the dynamic nature of the environment, which continuously interacts with and shapes human behaviors. Understanding this interplay is essential for devising effective cooperation strategies in SSDs, highlighting the need to balance behavior rules and immediate gains with long-term considerations.

\smallskip

{\bf (5)} {\em Revisiting existing theory of human cooperation,} particularly in light of recent findings indicating that AI systems, such as autonomous safety features in vehicles, can potentially disrupt established norms of reciprocity among humans~\citep{shirado2023emergence}. 
Traditionally, theories such as social norms~\citep{santos2018social}, prosocial preferences~\citep{fehr2003nature}, and reciprocity mechanisms~\citep{nowak2006five} have been pivotal in explaining cooperative behaviors among humans and other species. However, the emergence of a decline in reciprocity observed in semi-automatic driving~\citep{shirado2023emergence} prompts a deeper inquiry into the applicability of these theories within contexts where humans interact with agents. For instance, prosocial preferences suggest that individuals cooperate because they value the welfare of others; yet, it remains an open question whether such preferences extend to interactions involving non-human agents {\citep{shen2024beyond, burton2013prosocial,burton2016conditional}. Will individuals exhibit prosocial behaviors when their counterparts are perceived not as fellow humans but as agents? Can prosocial preferences still promote cooperation in environments where humans interact with AI? Similar inquiries are pertinent for other cooperative mechanisms, such as network reciprocity and both direct and indirect forms of reciprocity~\cite{rand2013human}. This evolving landscape calls for a critical reevaluation of the traditional theories of human cooperation, especially considering the growing integration of AI agents into human societies.

\section*{Acknowledgement}


This work was supported in part by the National Science Fund for the National Key R\&D Program (No. 2022YFE0112300), Distinguished Young Scholarship of China (No. 62025602), the National Natural Science Foundation of China (Nos. U22B2036, 11931015), Fok Ying-Tong Education Foundation China (No. 171105), the Fundamental Research Funds for the Central Universities (No. G2024WD0151), the Tencent Foundation and XPLORER PRIZE, China Postdoctoral Science Foundation (No. 2023M741852), JSPS Postdoctoral Fellowship Program for Foreign Researchers (No.~P21374), JSPS KAKENHI (No.~JP 22KF0303), and Shanghai Artificial Intelligence Laboratory.



\bibliographystyle{elsarticle-num-names} 
\bibliography{reference_short}





\end{document}